\documentclass[10pt,journal,compsoc]{IEEEtran}
\ifCLASSOPTIONcompsoc
  \usepackage[nocompress]{cite}
\else
  \usepackage{cite}
\fi
\ifCLASSINFOpdf
\else
\fi

\usepackage{amsmath}
\usepackage{multirow} 
\usepackage{multicol}
\usepackage{arydshln}
\usepackage{subfigure}
\usepackage{graphicx}
\usepackage[hyphens]{url}
\usepackage[numbers]{natbib}
\usepackage{ragged2e}
\usepackage{subfigure}
\usepackage{pgfplots}
\usepackage{booktabs}
\usepackage{hyperref}
\usepackage{longtable}
\usepackage[fontset=windows,scheme=plain]{ctex}

\begin{document}
\title{Unlikelihood Tuning on Negative Samples Amazingly Improves Zero-Shot Translation}
\author{
        Changtong Zan$^\dagger$,
        Liang Ding$^\ddagger$,
        Li Shen, 
        Yibin Lei,  
        Yibing Zhan,\\
        Weifeng Liu$^\ddagger$,~\IEEEmembership{Senior Member,~IEEE}
        and~Dacheng Tao,~\IEEEmembership{Fellow,~IEEE}
  \IEEEcompsocitemizethanks{
  \IEEEcompsocthanksitem C. Zan and W. Liu are with the College of Control Science and Engineering, China University of Petroleum (East China), Qingdao, China (e-mail: ctzan@s.upc.edu.cn; liuwf@upc.edu.cn).
  \IEEEcompsocthanksitem L. Ding is with the Shanghai Institute for Advanced Study, Zhejiang University, Shanghai, China (e-mail: liangding.liam@gmail.com)
  \IEEEcompsocthanksitem D. Tao is with the University of Sydney, Sydney, Australia (e-mail: dacheng.tao@gmail.com).
  \IEEEcompsocthanksitem L. Shen and Y. Zhan are with the JD Explore Academy at JD.com, Beijing, China (e-mail: mathshenli@gmail.com, zybjy@mail.ustc.edu.cn).
  \IEEEcompsocthanksitem Y. Lei is with the University of Amsterdam, Amsterdam, the Netherlands (e-mail: y.lei@uva.nl).
  \IEEEcompsocthanksitem 
  $^\ddagger$Corresponding Authors: Liang Ding (e-mail: liangding.liam@gmail.com) and Weifeng Liu (e-mail: liuwf@upc.edu.cn).
}
}

\IEEEtitleabstractindextext{%
\begin{abstract}
\justifying 
Zero-shot translation (ZST), which is generally based on a multilingual neural machine translation model, aims to translate between unseen language pairs in training data.
The common practice to guide the zero-shot language mapping during inference is to deliberately insert the source and target language IDs, e.g., $<$EN$>$ for English and $<$DE$>$ for German.
Recent studies have shown that language IDs sometimes fail to navigate the ZST task, making them suffer from the off-target problem (non-target language words exist in the generated translation) and, therefore, difficult to apply the current multilingual translation model to a broad range of zero-shot language scenarios.
To understand when and why the navigation capabilities of language IDs are weakened\footnote{We investigate and improve the target-side language ID.}, we compare two extreme decoder input cases in the ZST directions: \textit{Off-Target} (\textsc{Off}) and \textit{On-Target} (\textsc{On}) cases.
By contrastively visualizing the contextual word representations (CWRs) of these cases with teacher forcing,
we show that 1) the CWRs of different languages are effectively distributed in separate regions when the sentence and ID are matched (\textsc{On} setting), and 2) if the sentence and ID are unmatched (\textsc{Off} setting), the CWRs of different languages are chaotically distributed.
Our analyses suggest that although they work well in ideal \textsc{On} settings, language IDs become fragile and lose their navigation ability when faced with off-target tokens, which commonly exist during inference but are rare in training scenarios.
In response, we employ unlikelihood tuning on the negative (\textsc{Off}) samples to minimize their probability such that the language IDs can discriminate between the on- and off-target tokens during training.
Experiments conducted on the IWSLT, OPUS-100 (v1.0), WMT-5, and TED benchmarks spanning 40 ZST directions show that our method reduces the \textbf{off-target ratio} by \textbf{-48.0\%} on average, leading to a \textbf{+9.1 bilingual evaluation understudy (BLEU)} improvement with only an extra \textbf{+0.3\% tuning cost} on WMT-5. 
To facilitate reproducibility, we will publicly release our code at \url{github.com/zanchangtong/UNIONS}.
\end{abstract}

}

\maketitle

\IEEEdisplaynontitleabstractindextext

\IEEEpeerreviewmaketitle

\IEEEraisesectionheading{\section{Introduction}\label{sec:introduction}}

\begin{figure}[t] 
    \centering
    \includegraphics[width=0.51\textwidth]{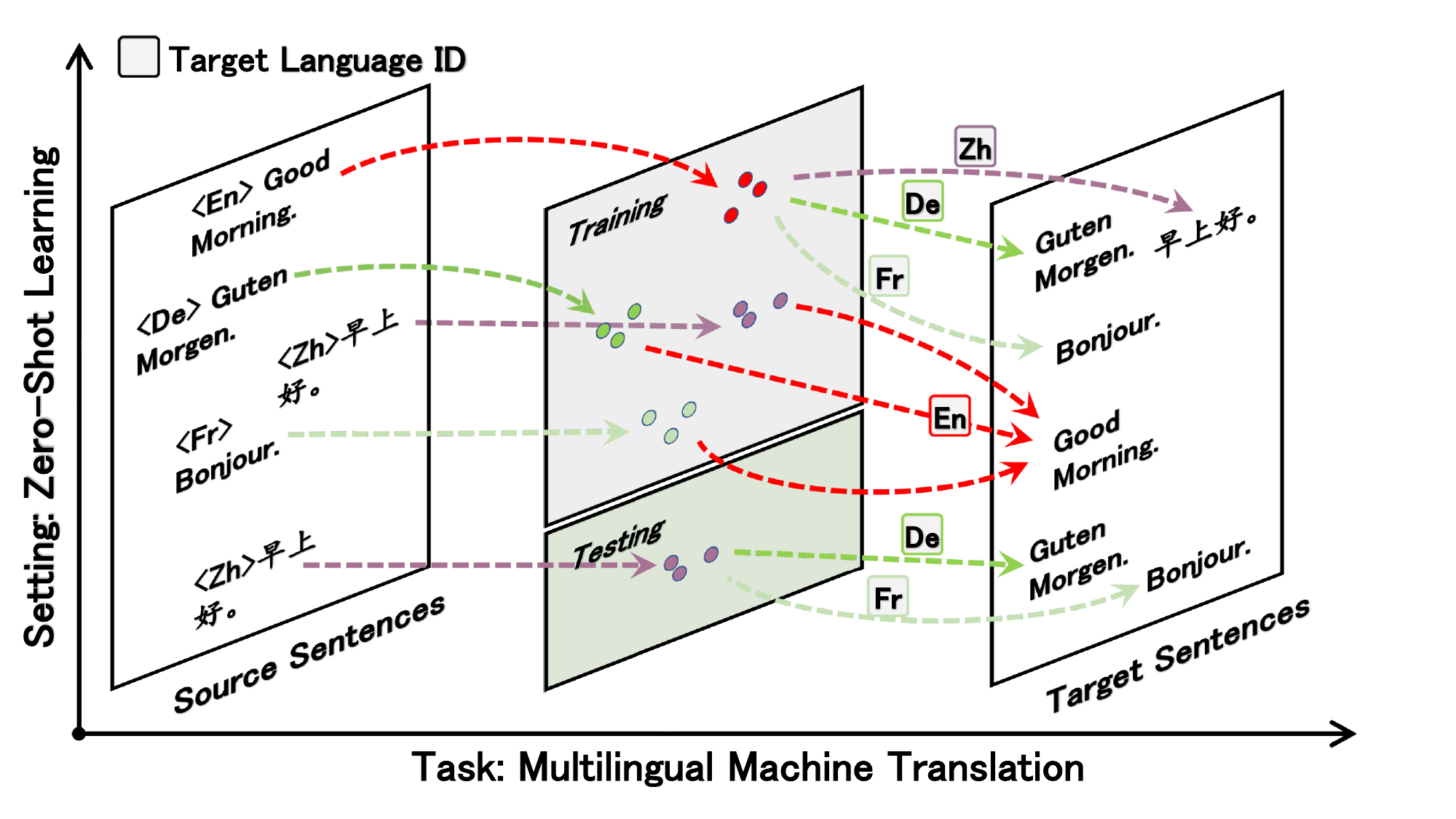}
    \caption{Zero-shot translation (\textbf{ZST}) aims to transfer the navigation ability of the target language ID into translation directions that do not exist in the training process.
    }
    \label{fig: zst_demo}
\end{figure}

\IEEEPARstart{M}{achine} translation (MT)~\cite{lopez2008statistical,DBLP:journals/corr/BahdanauCB14,DBLP:conf/ssst/ChoMBB14,ding2020understanding,ding-etal-2021-rejuvenating,ding-etal-2022-redistributing,zan2022complementarity,zan-etal-2022-vega,peng-etal-2023-token} has had a profound impact on people worldwide, revolutionizing various fields such as communication, writing, and travel, among others. However, the growth of machine translation has not been equally beneficial for all languages, particularly for low-resource languages such as Zulu, Yoruba, and Uzbek, which lack the necessary quantity of training data in each translation direction. As the number of languages increases, ensuring the availability of large-scale supervised data for training models becomes increasingly impractical due to the quadratic growth exhibited by the number of translation directions.
Consequently, zero-shot translation (ZST)~\cite{arivazhagan2019missing,gu-etal-2019-improved,zhang-etal-2020-improving,wang2022understanding,jin-xiong-2022-informative} has emerged as an intriguing solution, garnering attention from the research community. ZST leverages a unified multilingual neural machine translation (MNMT) model~\cite{dong-etal-2015-multi,firat-etal-2016-multi,firat-etal-2016-zero,johnson-etal-2017-googles,blackwood-etal-2018-multilingual,tan2018multilingual,aharoni-etal-2019-massively,arivazhagan2019massively} to translate between language pairs that are absent from the training dataset, as depicted in Figure~\ref{fig: zst_demo}.

Due to the transferability of knowledge between languages, an MNMT~\cite{zhang-etal-2020-improving,dabre2020survey,arthur-etal-2021-multilingual,ustun-etal-2021-multilingual,sun-etal-2022-alternative} model trained on multiple languages with a shared encoder-decoder framework possesses some degree of ZST ability, which is particularly useful for low-resource languages. However, the ZST task is still challenging since it requires one model to complete thousands of translation tasks with the guidance of the target language ID.

Recent studies~\cite{zhang-etal-2020-improving, gu-etal-2019-improved} have highlighted the primary obstacle to achieving satisfactory ZST performance: the \textit{off-target problem}. This issue arises when the language ID fails to effectively guide the associated model, leading to the inclusion of off-target tokens in the translation process. In our experiments, this problem is prevalent, with an incidence rate as high as \textbf{99.5\%}.
One line of research attributes this issue to the \textit{spurious correlations}~\cite{gu-etal-2019-improved, wang-etal-2021-rethinking-zero} between source languages and specific target languages. Such correlations arise due to the common practice of training models to translate all noncentral languages into a single central language (typically English).
Another perspective suggests that the problem lies in the \textit{missing ingredient}~\cite{arivazhagan2019missing, gu2022improving}, which fails to map different languages into a universal representation space during training. This absence of proper mappings confuses the decoding process and affects lexical choices.
However, most of these studies overlook the fact that each translated word is predicted based on the context words contained in decoder input, which are derived from the same language during training but may differ during inference. This discrepancy stems from the exposure bias between teacher forcing-based training paradigms and autoregressive inference.

In this paper, we revisit the off-target problem by exploring the gap between MNMT training and ZST inference.
It is well known that a translation model generates sequentially conditioned outputs on context words, which are ground-truth words during training but previous model outputs during inference~\cite{zhang-etal-2019-bridging}.
This gap inevitably leads the constructed MNMT model to predict ZST results conditioned on both correct (on-target) and erroneous (off-target) decodings during inference.
For further analysis, we focus on the contextual word representations (CWRs) output by the decoder and consider two extreme cases according to whether the language \& ID are matched: \textit{On-Target} (\textsc{On}) and \textit{Off-Target} (\textsc{Off}).
Interestingly, the CWRs of these two cases behave differently in the supervised and ZST directions.
1) In the \textsc{On} case, the CWRs are nicely distributed in separate regions based on the language ID, even in the ZST directions.
2) In the \textsc{Off} case, different languages are still distributed in separate regions in supervised directions, but they are chaotically distributed in the ZST directions.
In summary, the language ID loses its navigation ability only when the ID and decoder input are unmatched in ZST.
Therefore, we argue that the \textit{language ID has the potential to navigate arbitrary translation directions but becomes fragile when facing off-target decoder inputs due to the gap between training and inference.}

To address this issue, we propose an unlikelihood tuning method with \textit{inexpensive-to-obtain} negative (language- and ID-mismatched) samples by minimizing the probability of the negative samples. In this way, the language ID is encouraged to discriminate between the on/off-target tokens, whose capacity is hard to cultivate in the vanilla MNMT training procedure due to the lack of off-target samples but is required during ZST inference.
To construct the negative samples, we replace the target language ID in each translation pair with an ID that is randomly sampled from all languages except the current source and target languages.
We optimize the unlikelihood objective on these negative samples to minimize their probability and simultaneously optimize the MNMT objective on the (positive) translation samples to maintain the supervised translation ability. Additionally, as we cannot access the ZST samples during training, we select the model according to the degree of separation between the CWR distributions.

Experimentally, our method continued training upon the existing baseline methods\footnote{To ensure effective performance, the baseline models are trained with their optimal training scripts.} substantially reduces the off-target ratio by -16.7\%, -79.2\%, -60.0\%, and -14.8\% on the IWSLT-4, OPUS-100 (v1.0), WMT-5, and TED benchmarks, respectively, thus significantly improving the resulting translation quality by +4.0, +14.6, +9.1, and +1.3 SacreBLEU points.
The main \textbf{contributions} are as follows.
\begin{itemize} 
    \item We show that language IDs have the potential to navigate arbitrary translation directions but become fragile when facing off-target decoder inputs due to the gap between the MNMT training and ZST inference processes. 
    \item We propose a method to minimize the probability of negative (language- and ID-mismatched) samples with unlikelihood tuning for the pretrained MNMT models.
    \item Our \textit{simple-but-sufficient} method exhibits significant and consistent ZST improvements (up to 20.8 bilingual evaluation understudy (BLEU) by reducing the off-target ratio by 88.2\%) over the baseline while maintaining its supervised translation quality.
\end{itemize}

The rest of this paper is organized as follows. \S\ref{sec: related works} discusses some related works. \S\ref{sec: background} introduces the background of the study. In \S\ref{sec: gap}, we revisit the gap between the MNMT training and ZST inference processes and reveal the impact of off-target tokens. Our approach, \textbf{UN}l\textbf{I}kelihood tuning \textbf{O}n \textbf{N}egative \textbf{S}amples (UNIONS), is presented in \S\ref{sec: unions}. The experimental setup is demonstrated in \S\ref{sec: task}. The main experimental results are reported in \S\ref{sec: main results}, followed by analysis experiments in \S\ref{sec: analysis}. Finally, conclusions are presented in \S\ref{sec: conclusion}.

\section{Related Works}
\label{sec: related works}
\subsection{Multilingual Neural Machine Translation}
MNMT~\cite{dong-etal-2015-multi,firat-etal-2016-multi,firat-etal-2016-zero,johnson-etal-2017-googles,blackwood-etal-2018-multilingual,tan2018multilingual,aharoni-etal-2019-massively,zhang-etal-2020-improving,arthur-etal-2021-multilingual,ustun-etal-2021-multilingual,ding2022improving,sun-etal-2022-alternative} aims to facilitate translation between multiple languages using a single model, typically by employing a designated token to indicate the target language for a given source sentence.
\citet{johnson-etal-2017-googles} introduced utilizing an artificial token at the beginning of the input sentence to specify the desired target language and enable the use of multilingual translation systems with a single model.
\citet{blackwood-etal-2018-multilingual} proposed a task-specific attention model, which demonstrated improvements over a fully shared model in terms of translation quality.
\citet{aharoni-etal-2019-massively} took a significant step toward developing a massively multilingual translation model by expanding the number of languages supported by MNMT to 102.
\citet{zhang-etal-2020-improving} further enhanced the translation performance of their massively multilingual translation model by leveraging larger model capacities. They also discovered that language-specific modeling and deep architectures improve ZST, albeit falling short in terms of addressing the off-target problem.
\citet{ustun-etal-2021-multilingual} explored the unsupervised setting, where the translation is performed between languages with only monolingual data, by incorporating denoising adapters on top of pretrained language models.
Recently, \citet{sun-etal-2022-alternative} introduced alternative signals, such as phonetic, romanized, and transliterated inputs, to MNMT to enhance the transferability of the constructed model across different languages.

However, these methods are not specifically tailored for the ZST task. In contrast, our method focuses on enhancing the navigation ability of the target language ID and can serve as a complementary plug-in algorithm to augment these existing approaches. This consideration will be further explored in our subsequent studies.

\subsection{MNMT-Based ZST}
The existing works have mainly focused on two perspectives to mitigate the off-target translation problem.
1) Introducing more inductive regularizers. \citet{arivazhagan2019missing} forced their encoder to present different sentences with language-invariant representations to improve its generalization ability.
\citet{jin-xiong-2022-informative} found that adding more target language information has a significant impact on translation performance, especially for ZST.
\citet{wang2022understanding} focused on the uncertainty property and proposed the use of data cleaning and vocabulary masking to reduce the off-target ratio.

2) Introducing more positive samples, e.g., non-English sentence pairs. \citet{gu-etal-2019-improved} noted that the spurious correlation issue leads to inferior zero-shot performance, and they constructed positive paired samples offline, which is a similar approach to that of \citet{JMLR:v22:20-1307}. In addition, \citet{zhang-etal-2020-improving} found that the synthetic parallel data generated by online self-training also help with ZST.
However, it is difficult to construct positive datasets for the more than 7100*7100 language pairs\footnote{\url{https://en.wikipedia.org/wiki/Lists_of_languages}} in the world.

In contrast, we reveal the gap between the MNMT training and ZST inference processes and propose a simple and effective method to bridge this gap, which has the potential to complement the existing approaches with a simple strategy: performing continued training with our method on their existing checkpoints.
Although \citet{pan-etal-2021-contrastive} also used negative samples, their method completely differs from ours, as they employed contrastive learning on output encoder representations while training from scratch, whereas we address the training-inference gap by performing simple unlikelihood tuning.

\subsection{Exposure Bias of Sequence Generation}
Exposure bias~\cite{ranzato2016sequence} refers to the discrepancy between the training and inference processes in sequence generation tasks. During training, models are conditioned on ground-truth tokens, whereas during inference, they predict tokens based on their previous outputs.
Numerous methods have been developed to address this issue. Scheduled sampling, proposed by \citet{bengio2015scheduled}, gradually replaces the ground-truth tokens with tokens predicted by the model itself. To alleviate overcorrection and noise perturbations, \citet{zhang-etal-2019-bridging} introduced a sentence-level oracle, improving the predicted distribution.

Another approach involves training models using non-maximum likelihood estimation (non-MLE) objectives to mitigate exposure bias. Minimum risk training (MRT), proposed by \citet{shen-etal-2016-minimum}, directly optimizes the model parameters with respect to arbitrary evaluation metrics. \citet{bahdanau2017an} employed an actor-critic algorithm from the reinforcement learning (RL) domain to train models for sequence generation, enabling the training procedure to approximate the testing process and directly optimize the evaluation scores. Additionally, \citet{du-ji-2019-empirical} combined RL-based imitation learning with a pointer-generator framework as their base model.

Several studies have focused on analyzing exposure bias. \citet{wang-sennrich-2020-exposure} and \citet{schmidt-2019-generalization} established a link between exposure bias and the generation gap resulting from distribution and domain shifts. \citet{he-etal-2021-exposure} discovered the self-recovery ability possessed by language models, which can counteract the harmful effects of exposure bias. From the perspective of imitation learning, \citet{arora-etal-2022-exposure} connected the degeneration problem encountered by large language models to exposure bias.

However, most existing methods primarily address supervised tasks. In our work, we identify a specific type of exposure bias in ZST that leads to the off-target problem. We aim to alleviate the misleading of model outputs in the wrong language.

\subsection{Unlikelihood Training}
To address the degeneration problem encountered during neural language generation, \citet{welleck2019neural} highlighted the drawback of the likelihood loss itself and first proposed a complementary unlikelihood training objective, which forces unlikely samples to be assigned lower probabilities by the model.
This method has been further explored in dialog tasks by \citet{li-etal-2020-dont}, who demonstrated its effectiveness in generating more consistent and coherent human-like dialog. Moreover, \citet{song-etal-2021-bob} proposed a transformers (BERT)-based dialog model and demonstrated the benefits of incorporating unlikelihood training with nondialogue inference data to enhance the understanding capabilities of the resulting model.
Additionally, \citet{nogueira-dos-santos-etal-2020-beyond} used the unlikelihood loss for ranking and proposed a generative information retrieval approach.
\citet{hosseini-etal-2021-understanding} proposed the combination of an unlikelihood objective with a reference-based setup for input sentences to model negation with pretrained BERT~\cite{devlin-etal-2019-bert}.
Recently, \citet{10.1145/3503161.3547974} proposed using unlikelihood training on a visual dialog model to reduce the probability of producing wrong answers and achieve state-of-the-art performance.

In this work, we focus on the ZST task and propose a new unlikelihood training method for our constructed negative translation samples to reduce the probability of off-target tokens, thus alleviating the severe off-target problem.

\section{Background}
\label{sec: background}
In this section, we present the standard frameworks for neural machine translation (NMT) and MNMT models.

\subsection{Neural Machine Translation}
NMT is a field that focuses on converting sentences from a source language into target language representations using neural networks. Initially, the early works in NMT primarily utilized recurrent neural networks (RNNs) as their foundational models. However, in recent years, encoder-decoder transformers~\cite{NIPS2017_3f5ee243} have emerged as the dominant architectures due to their superior parallelization capabilities and impressive performance. Both RNN and transformer models heavily rely on the teacher forcing-based training method and the autoregressive inference paradigm to effectively implement a translation system.

During the training stage of NMT, given a sample $S_i = \left ( X_i, Y_i \right ) = \left ({x}_1^{i},..,{x}_n^{i}; {y}_1^{i},..,{y}_m^{i}\right )$, the encoder $ENC$ maps the source sentence into the latent feature space to obtain $H^{i}$. Then, the decoder predicts each token based on both the encoder output $H^{i}$ and the previous tokens $y_{<t}^{i}$ in the target sentence as follows:
\begin{eqnarray}
\label{eq:NMT predict token}
\mathbf{H}^{i} &=& \textit{ENC} \left( {x}_1^{i},..,{x}_n^{i} \right) \nonumber \\
\mathbf{P}\left( y_t^{i} | x^{i}, y_{<t}^{i}\right) &=& \textit{DEC} \left( H^{i}, y_{<t}^{i} \right ) \nonumber 
\end{eqnarray}
where $t$ represents the position of a token in the sentence, and $P$ denotes the model output probability of the target token. During the training process of the NMT model, we provide the preceding ground-truth tokens $y_{<t}^{i}$ to assist with predicting the current token. The training objective is to maximize the log-likelihood of the training data with the following loss function:
\begin{eqnarray}
\label{eq:NMT loss}
\mathbf{L}_{\textit{ENC, DEC}} = & - \sum_{i=1}^{N} \sum_{t=1}^{m} log \left( {P}\left( y_t^{i} | x^{i}, y_{<t}^{i}\right) \right) 
\nonumber
\end{eqnarray}

During the inference stage, our objective is to translate sentences into the target language representation without having access to any ground-truth tokens. The autoregressive inference paradigm utilizes the model to predict each token individually. Specifically, we replace the ground-truth tokens $y_{<t}^{i}$ with the model-generated tokens from the previous steps and continue the loop until a stop word appears.

\subsection{Multilingual Neural Machine Translation}
Following \citet{johnson-etal-2017-googles}, the objective of MNMT is to train a model that can handle multiple translation tasks by incorporating language IDs to provide translation direction guidance. To be more specific, the prediction process of MNMT in one direction can be formulated as follows. Given a source language ID $l_s$ and a target language ID $l_t$, the predicted probability distribution of the $t$-th target token in our encoder-decoder-based translation system is:
\begin{eqnarray}
\label{eq:off-target_tokens}
\mathbf{H}^{i} &=& \textit{ENC} \left( l_s, {x}_1^{i},..,{x}_n^{i} \right) \nonumber \\
\mathbf{P}\left( y_t^{i} | x^{i}, y_{<t}^{i}, l_t\right) &=& \textit{DEC} \left( H^{i}, y_{<t}^{i}, l_t \right ) \nonumber 
\end{eqnarray}
where the target language ID $l_t$ and the decoder input $y_{<t}$ work together to determine which language should to translate into. Compared with bilingual translation, we jointly optimize multiple translation tasks and use language IDs to navigate the translation process, which is more efficient for translating between many languages.
Furthermore, the MNMT model can translate between language pairs that are not present in the training data by appropriately setting the IDs in the ZST directions~\cite{johnson-etal-2017-googles}.

\section{Rethinking the Off-Target Problem}
\label{sec: gap}
In this section, we first introduce our MNMT model and highlight the disparity between training it with MNMT data and inferring it with ZST tasks. Next, we delve into an analysis of when and why the MNMT model produces off-target translations, specifically focusing on the role of CWRs and providing insights into the reasons behind these deviations.

\subsection{Multilingual Machine Translation Model}
We first present the key settings used to train the base MNMT model on the IWSLT-4 dataset for exploratory analysis purposes. As IWSLT-4 is a multialigned dataset, we concatenate sentences from all languages to train a 40k SentencePiece~\cite{kudo-richardson-2018-sentencepiece} vocabulary and then use it to tokenize all data into subword units.
To distinguish between different translation tasks, we prepend the corresponding language IDs to the source and target sentences. More detailed settings are presented in \S\ref{sec:RI_settings}.

\subsection{Gap Between MNMT Training and ZST Inference}
As mentioned by \citet{ranzato2016sequence}, an NMT model is typically trained to predict the next token in a sequence given the previous tokens. However, during the testing phase, the model is supposed to generate the entire sequence from the beginning. This discrepancy, commonly known as exposure bias, can make the generation process fragile due to the accumulation of errors.

In this paper, we establish a connection between exposure bias and ZST. We show that the gap between the MNMT training and ZST testing processes is a crucial cause of the terrible off-target problem. To elaborate, we present the formulation of ZST as follows:
\begin{eqnarray}
\label{eq:off-target_tokens}
\mathbf{P}_{t} = \textit{DEC} \left( \textit{ENC}\left ( X, l_s \right ) , l_t, P_{<t} \right ) 
\end{eqnarray}
where the translation direction $l_s \rightarrow l_t$ does not exist during MNMT training and $P_{<t}$ denotes the previous model outputs.
During MNMT training, the teacher forcing paradigm forces the decoder input $P_{<t}$ to be the ground-truth tokens in the correct language $l_t$.
During the inference stage, this condition is broken~\cite{zhang-etal-2019-bridging}.
Autoregressive inference predicts the target words based on the previous model outputs, and error tokens $P_{<t}$ (we only consider the off-target tokens that are present in another language rather than $l_t$) commonly exist.

\begin{figure*}[!htb]
\centering
\subfigure{
\begin{minipage}[b]{0.48\linewidth}%
\setlength{\abovecaptionskip}{0pt}
\begin{center}
     \includegraphics[width=1.0\linewidth]{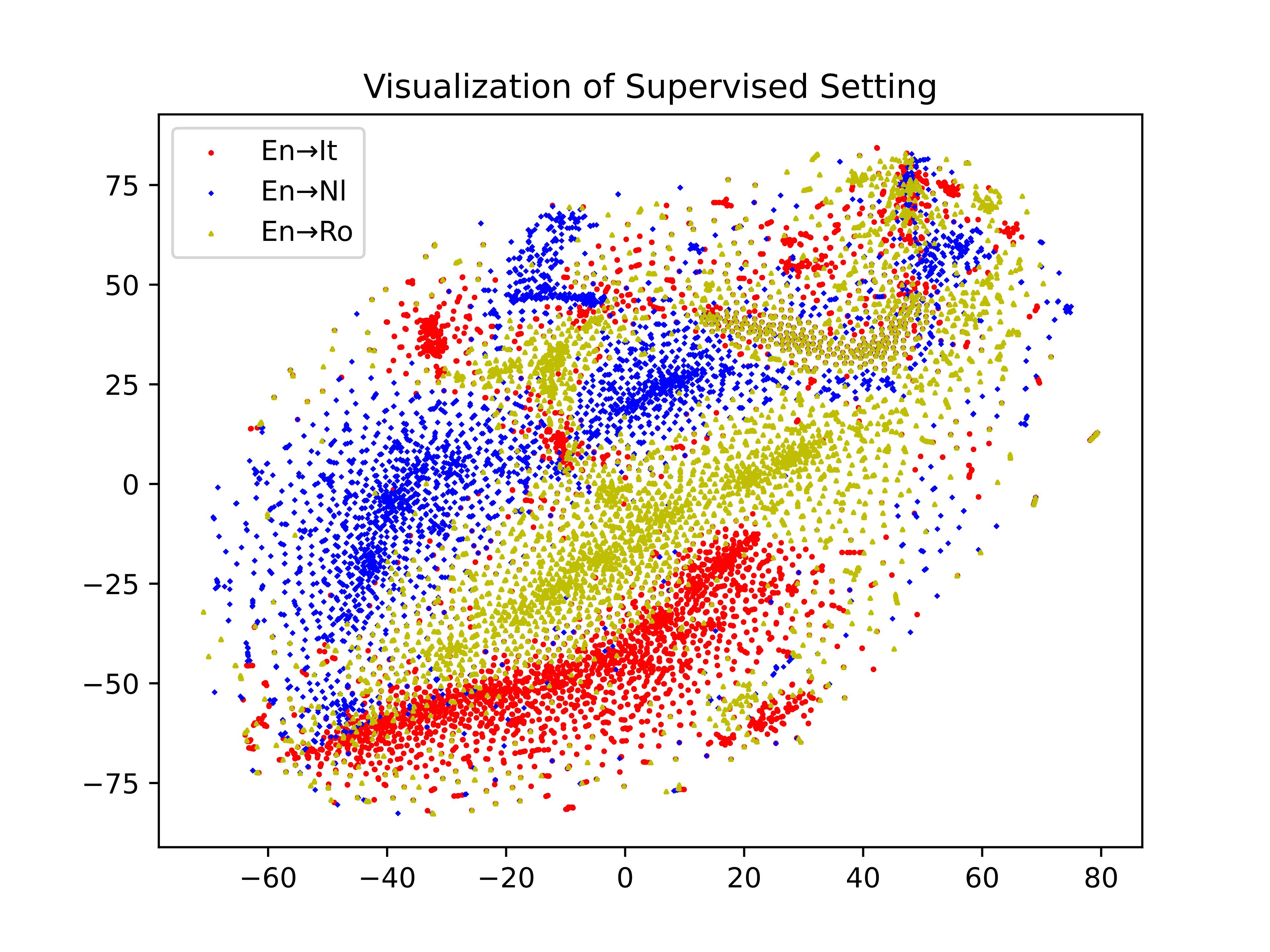}
     \centerline{(a) Supervised on-target case}
\end{center}
\end{minipage}
}
\subfigure{
\begin{minipage}[b]{0.48\linewidth}%
\setlength{\abovecaptionskip}{0pt}
\begin{center}
     \includegraphics[width=1.0\linewidth]{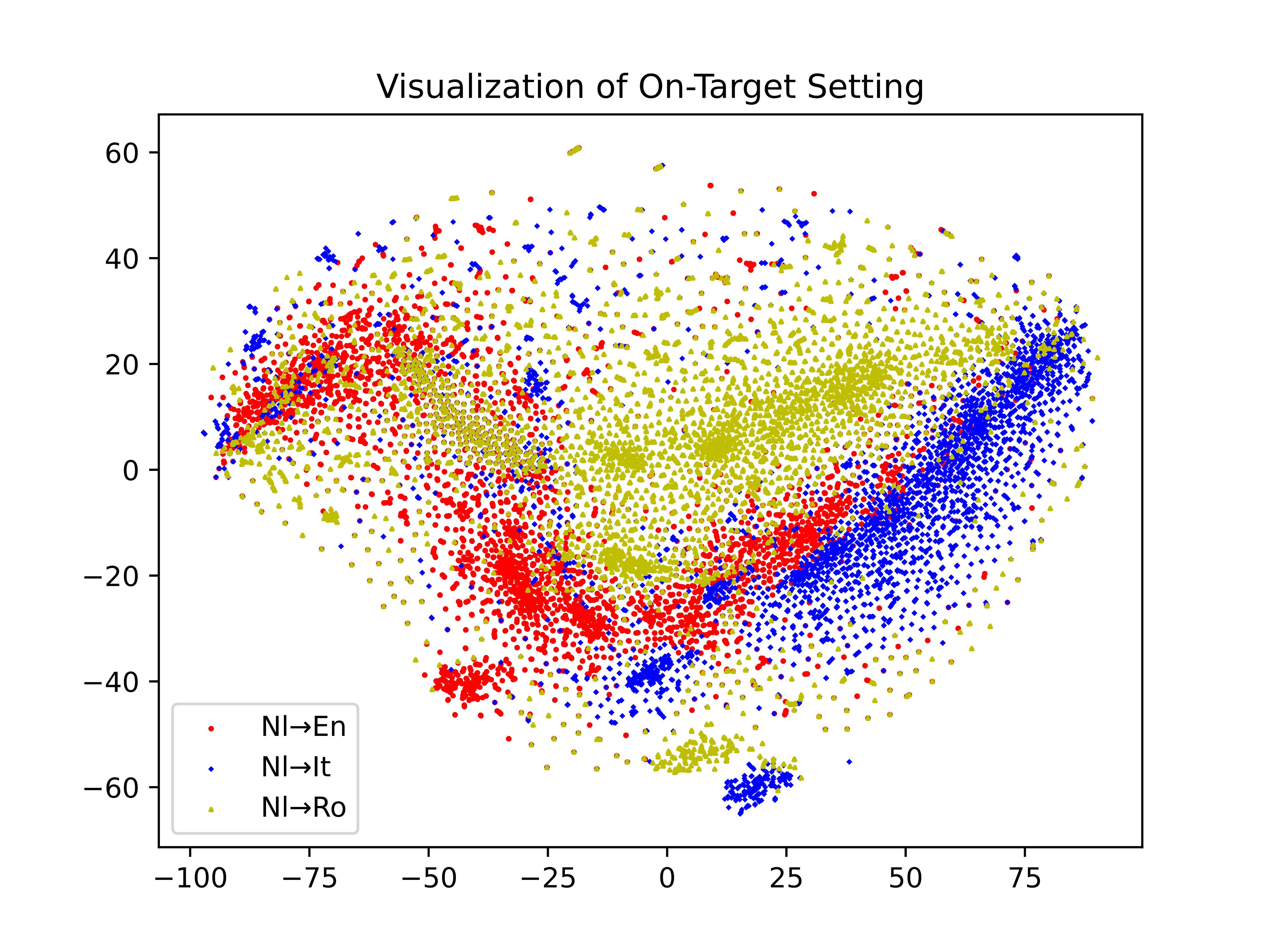}
     \centerline{(b) Zero-shot on-target case}
\end{center}
\end{minipage}
}
\subfigure{
\begin{minipage}[b]{0.48\linewidth}%
\setlength{\abovecaptionskip}{0pt}
\begin{center}
     \includegraphics[width=1.0\linewidth]{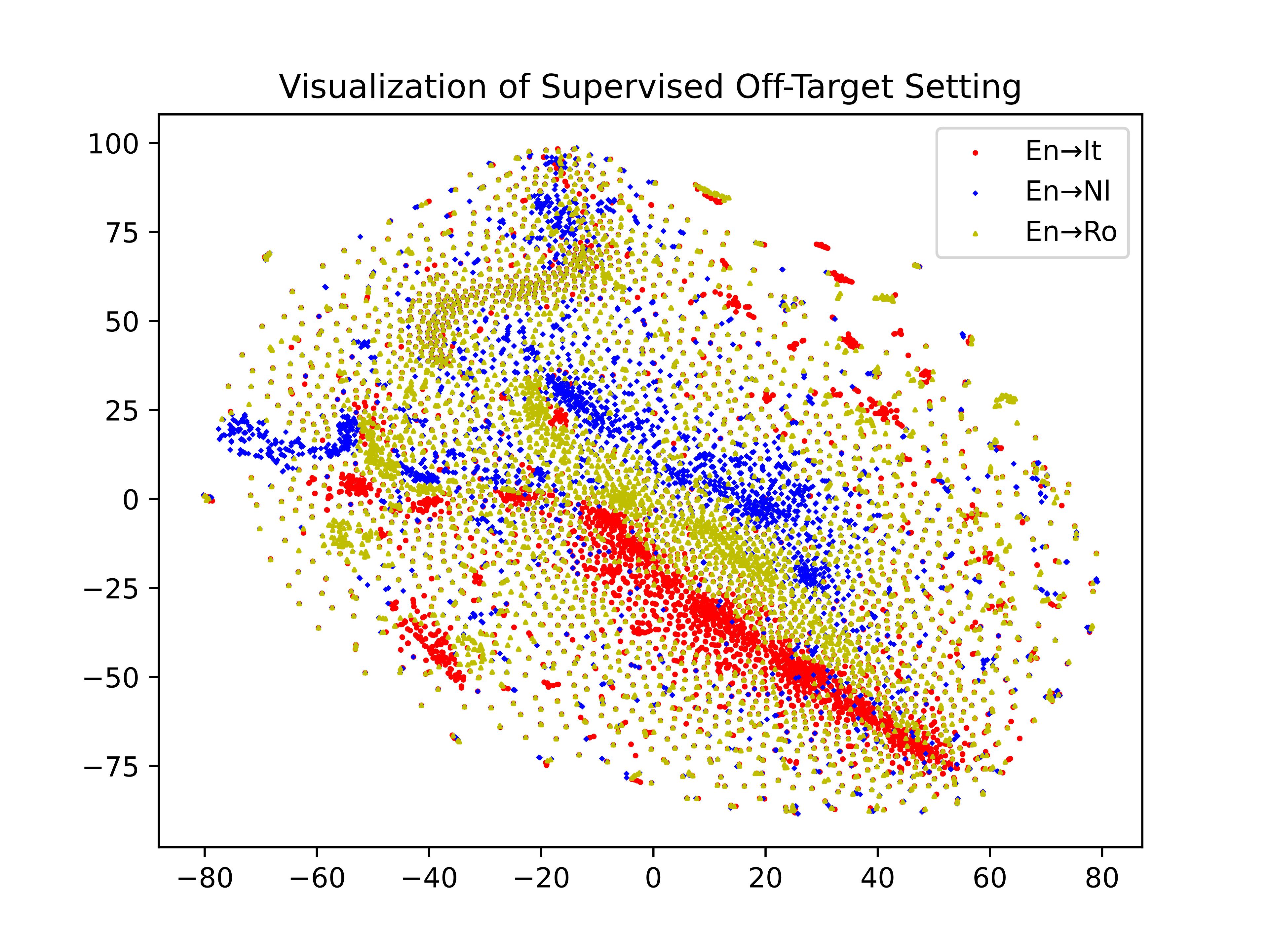}
     \centerline{(c) Supervised off-target case}
\end{center}
\end{minipage}
}
\subfigure{
\begin{minipage}[b]{0.48\linewidth}%
\setlength{\abovecaptionskip}{0pt}
\begin{center}
     \includegraphics[width=1.0\linewidth]{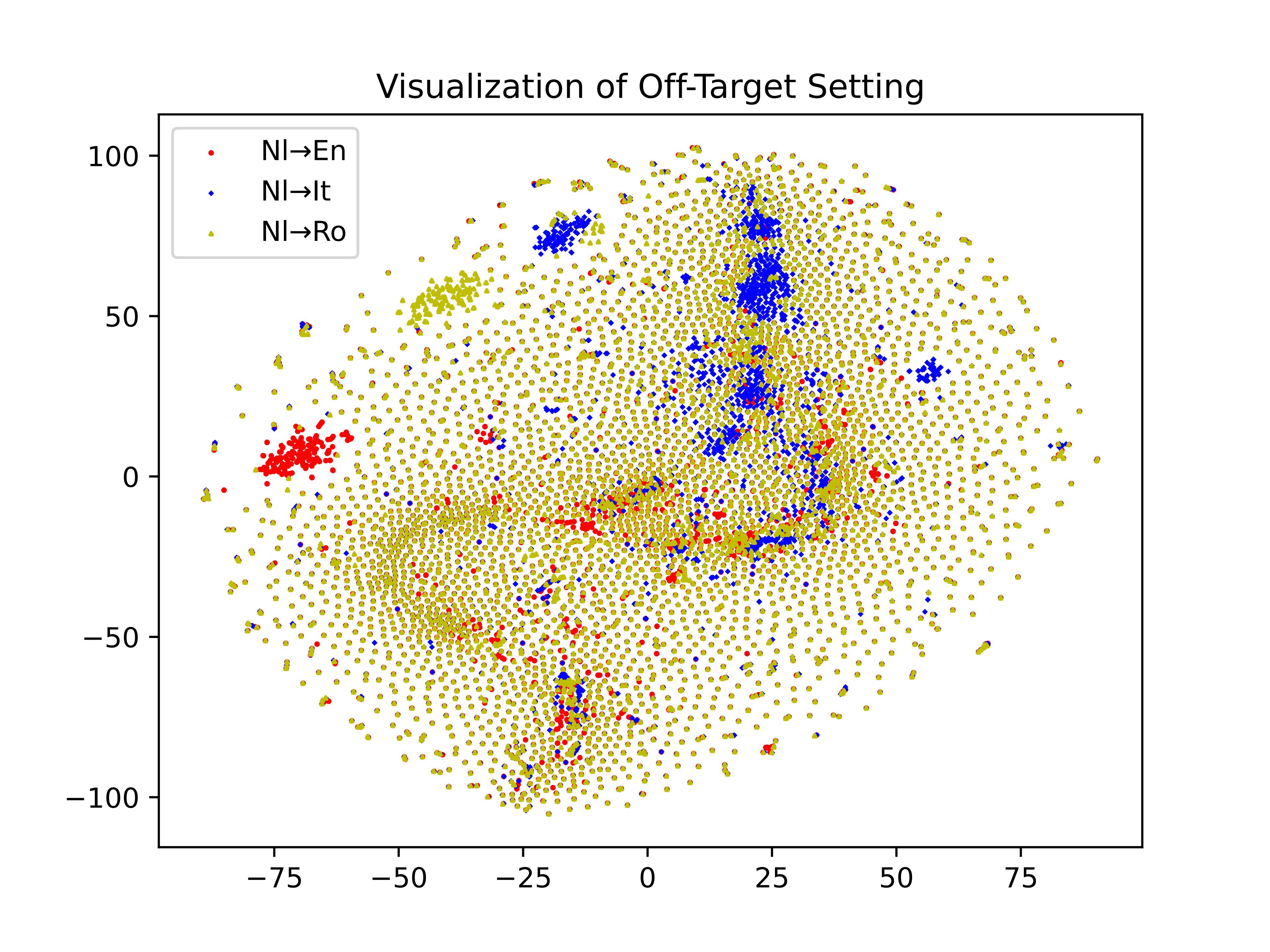}
     \centerline{(d) Zero-shot off-target case}
\end{center}
\end{minipage}
}

\caption{\textbf{Comparative visualization among different CWRs} on the IWSLT-4 dataset. The on-target settings in (a) and (b) are distributed in separate regions, which means that the language ID can navigate the translation flow. Contrary to the supervised off-target setting in (c), the MNMT model predicts the CWRs of a mixed distribution in (d) and shows that, with the distraction provided by the off-target tokens, the navigation ability of the language ID is covered by ZST. }
\label{fig:cwr_vis}
\end{figure*}

\subsection{Impact of Off-Target Tokens}
\label{three_setting_for_vis}
To understand how off-target tokens in the decoder input affect the translation flow, we focus on the output CWR of the decoder, which is computed with Equation~\ref{eq:off-target_tokens}, and consider two extreme cases, the \textsc{on}-target and \textsc{off}-target cases, in supervised translation and ZST.
\begin{itemize} 
    \item \textbf{Supervised on-target setting}: CWRs of supervised directions (En$\rightarrow$XX) with ground-truth decoder input. 
    \item \textbf{Zero-shot on-target setting}: CWRs of zero-shot directions (Nl$\rightarrow$XX) with ground-truth decoder input. 
    \item \textbf{Supervised off-target setting}: CWRs of supervised directions (En$\rightarrow$XX), decoder input is in Nl with the same content as source sentence. 
    \item \textbf{Zero-shot off-target setting}: CWRs of zero-shot directions (Nl$\rightarrow$XX), decoder input is in En with the same content as source sentence. 
\end{itemize}

Notably, we include Nl$\rightarrow$En, which is available in the training data, in both zero-shot settings for comparison, as the off-target translation is usually done in English. To address the language coverage bias~\cite{wang-etal-2021-language-coverage}, we sampled 200 sentences with the same content. We perform visualization with t-distributed stochastic neighbor embedding (t-SNE)~\cite{van2008visualizing}.

First, we compare \textbf{on-target settings} in which the language ID and decoder input are matched. As shown in Figure~\ref{fig:cwr_vis}(a), with the matched language ID and decoder input in the supervised directions, the CWRs are well grouped in separate regions according to their languages, which may be due to the MNMT model that translates English sentences is unlikely to predict off-target words, which is consistent with previous work~\cite{yang-etal-2021-improving-multilingual}. For the zero-shot directions, in Figure~\ref{fig:cwr_vis}(b), we can see a similar phenomenon with the CWR distributions in the supervised directions. This suggests that the \textit{language ID is able to navigate an arbitrary translation flow into the correct language under the ideal on-target setting}.

We next set all ZST decoder inputs to be off-target tokens (\textbf{off-target setting}), as presented in Figures~\ref{fig:cwr_vis}(c) and (d). Contrary to the phenomenon observed in the supervised directions (c), the CWRs in the zero-shot directions (d) are chaotically distributed, which means that the MNMT model does not know which language the text should be translated into.
Therefore, we argue that the \textit{language ID tends to be fragile and loses its navigation ability when faced with off-target tokens.}

\begin{figure*}[t] 
    \centering
    \includegraphics[width=0.91\textwidth]{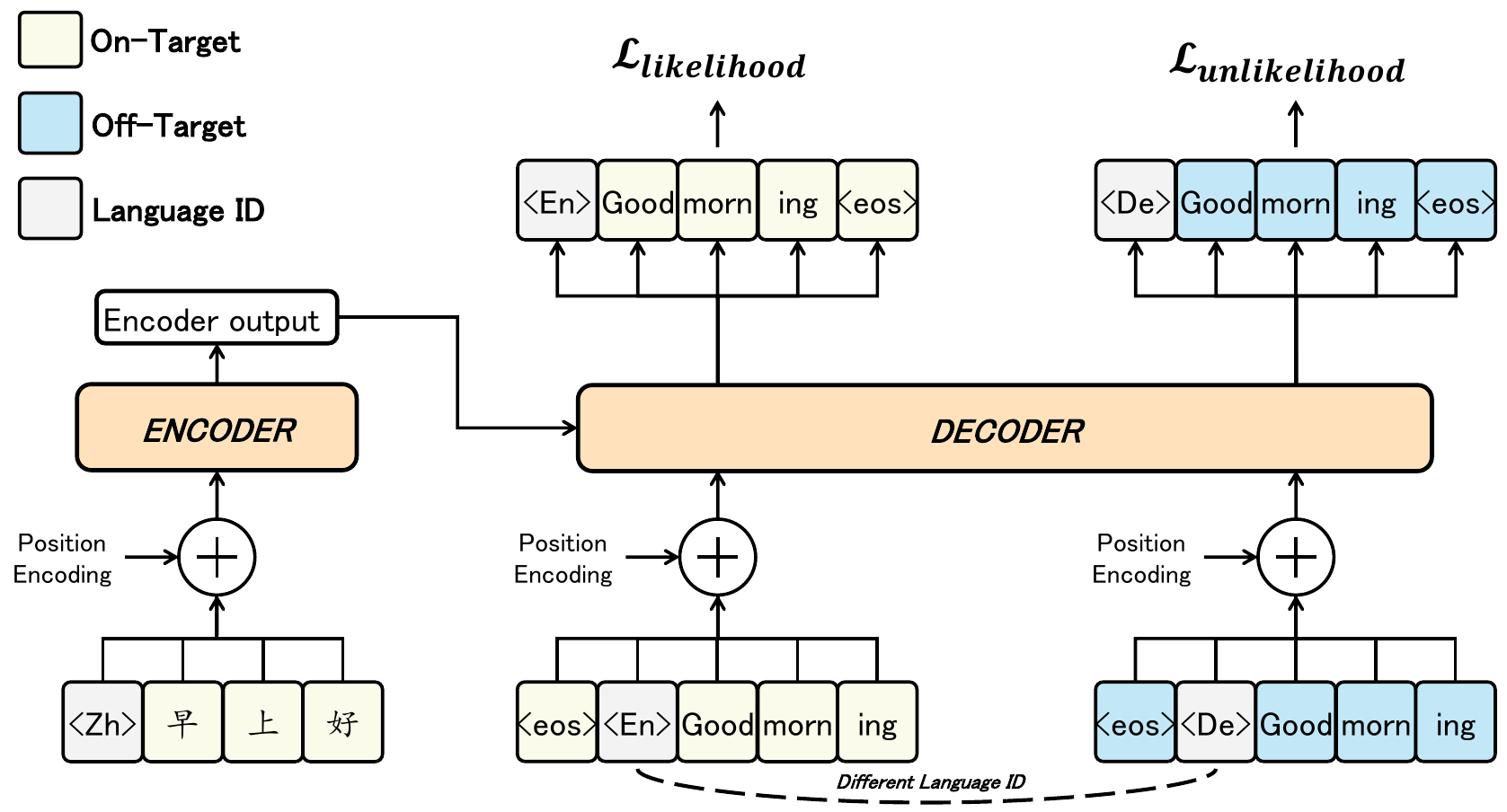}
    \caption{
    \textbf{The training scheme of UNIONS}. Given an on-target translation pair, we feed the source/target sentences into the encoder/decoder with the corresponding IDs to maximize the likelihood loss objective, i.e., $\mathcal{L}_{likelihood}$. For negative samples whose only difference from the translation samples is the off-target language ID, we minimize the unlikelihood loss, i.e., $\mathcal{L}_{likelihood}$. 
    }
    \label{fig: main figure}
\end{figure*}

\section{Methodology}
\label{sec: unions}
In this section, we introduce the problem definition and present our proposed method, \textbf{UN}l\textbf{I}kelihood tuning \textbf{O}n \textbf{N}egative \textbf{S}amples (UNIONS).

\subsection{Problem Definition}
Given an MNMT training dataset $D = \left(D_{l_1 \leftrightarrow En},..., D_{l_N \leftrightarrow En} \right)$ including languages $L= \{l_1,...,l_N, En \}$, and $D_{l_s \rightarrow l_t} = {\left ( X^{l_s}, Y^{l_t} \right )}$, the parallel data in $N*\left( N-1\right)$ ZST directions are only available for evaluation. While all languages exist in the training dataset, the goal of UNIONS is to train an MNMT model for an arbitrary ZST direction.  

\subsection{General Framework}
We design the UNIONS method with the goal of enhancing the ZST translation performance of a pretrained MNMT model. To achieve this, we first initialize the model with pretrained MNMT parameters. Next, we simultaneously minimize the likelihood loss on the supervised training data and the unlikelihood loss on the coupled negative samples. This approach helps address the off-target problem.

The general objective function is formulated as follows:
\begin{small} 
\begin{eqnarray}
\label{eq:general framework}
\mathcal{F} =& \mathop{\arg\min} \sum_{l=l_1}^{l_N}\mathcal{L}_{MLE} \left(X^{l}, Y^{En}, \theta \right) \!+\! \mathcal{L}_{MLE} \left(X^{En}, Y^{l}, \theta \right) \nonumber \\ 
& \!+\! \mathcal{L}_{UL} \left(X^{l}, \tilde{Y}^{En}, \theta \right) \!+\! \mathcal{L}_{UL} \left(X^{En}, \tilde{Y}^{l}, \theta \right)  \nonumber
\end{eqnarray}
\end{small} 
where $\theta$ denotes the parameters of the model, $\left( X, \tilde{Y} \right)$ represents the coupled negative samples of $\left( X, Y \right)$, $\mathcal{L}_{MLE}$ is the likelihood loss induced on the supervised training data, and $\mathcal{L}_{UL}$ represents the unlikelihood loss induced on our negative samples.

The commonly used early-stopping strategy relies on the validation loss induced on the training dataset. However, this strategy is not suitable for predicting ZST performance, as the dev set only comprises data in the supervised directions. It is possible for a model with a lower validation loss to achieve a worse ZST score. Hence, we propose an additional approach for selecting the final model based on the CWR separation degree in the zero-shot off-target setting.

The details of each part are discussed in the following subsections. 

\subsubsection{Likelihood Tuning on Supervised Samples}
We minimize the maximum likelihood estimation (MLE) loss on the supervised training data and aim to inherit the translation ability of the pretrained MNMT model. Specifically, we optimize multiple bilingual translation tasks on the multilingual translation dataset $\mathcal{D}$, which consists of En$\rightarrow$XX and XX$\rightarrow$En bilingual corpora. Given a sentence pair $\left( X_i, Y_i \right) =\left( x_1, x_2,..., x_n; y_1, y_2,..., y_m\right )$ with an index $i$ in $\mathcal{D}$, where $l_s, l_t$ are the source and target languages, respectively, $x, y$ denote the words of $X_i, Y_i$ with lengths of $n, m$, respectively. 

As shown in the center of Figure~\ref{fig: main figure}, we append the corresponding language IDs $\left \langle l_s \right \rangle, \left \langle l_t \right \rangle $ to the sentence pairs to construct positive translation samples, where the IDs distinguish different between translation tasks; e.g., the Zh-En sample can be denoted as ``\begin{CJK}{UTF8}{gbsn}$\langle$Zh$\rangle$ 早上好 \end{CJK}'' and ``$\langle$En$\rangle$ Good morning'' with ID tokens ``$\langle$Zh$\rangle$'' and ``$\langle$En$\rangle$'', respectively.
First, we feed the source sentence into the encoder to obtain the encoder output feature, which consists of token-level features. Based on the encoder output and the decoder input, the decoder predicts the probabilities of the current token.

Then, we optimize the encoder-decoder-based translation model, e.g., a transformer~\cite{NIPS2017_3f5ee243}, with the following likelihood loss objective:
\begin{small}
\begin{eqnarray}
\label{eq:mnmt}
\mathcal{L}_{MLE} \left(X^{l_s}, Y^{l_t}, \theta \right)  = & \nonumber \\
-\!\sum_{\left ( X_i^{l_s}, Y_i^{l_t}\right ) \in \mathcal{D}}\! \sum_{t \in m}\!&\log P(y_t | l_s, l_t, X_i, y_{<t}, \theta) \nonumber,
\end{eqnarray}
\end{small}
where $\left( X_i, Y_i \right)$ is the translation sentence pair for training and $y_{<t}$ denotes the input words of the decoder. Multilingual machine translation maximizes the probability of predicting correct tokens conditioned on the source sentence, the language IDs, and the ground-truth decoder inputs.

Considering the multiple directions contained in the training data, the final objective function of the likelihood training can be formulated as follows:
\begin{small}
\begin{eqnarray}
\label{eq:MNMT training objective}
\mathcal{F}_{MNMT}=& \nonumber \\
\mathop{\arg\min} \sum_{l=l_1}^{l_N}\mathcal{L}_{MLE}& \left(X^{l}, Y^{En}, \theta \right)\!+\! \mathcal{L}_{MLE} \left(X^{En}, Y^{l}, \theta \right) \nonumber 
\end{eqnarray}
\end{small}

\subsubsection{Unlikelihood Tuning on Negative Samples}
We also tune the pretrained MNMT model with negative samples to bridge the gap between the MNMT train and ZST inference processes. Specifically, as illustrated on the right side of Figure~\ref{fig: main figure}, for each translation sentence pair with language IDs $l_s, l_t$, we build a negative language ID set $L_{ne} = \left \{ l \in L, l \neq l_t ~\text{and}~ l \neq l_s \right \}$. During training, we randomly select a language ID $l_{ne}$ from $L_{ne}$ to replace $l_t$, i.e., ``$\langle$En$\rangle$'' $\rightarrow$ ``$\langle$De$\rangle$'', to construct a negative sample pair $\left(X^{l_s}, \tilde{Y}^{l_{ne}} \right)$. As they are not consistent with the target language ID $l_{ne}$, we call the tokens of the target sentence ``off-target tokens''.
The model is then trained on the constructed negative samples with the following unlikelihood loss:
\begin{small}
\begin{eqnarray}
\label{eq:unlikelihood_loss}
\mathcal{L}_{UL} \left(X^{l_s}, \tilde{Y}^{l_{ne}} , \theta \right) \!= & \nonumber \\
-\! \sum_{\left(X^{l_s}, \tilde{Y}^{l_{ne}} \right) \in \mathcal{D}} \sum_{t \in m} &\log \left(1 \!-\! P(\tilde{y}_t | l_s, l_{ne}, X_i, \tilde{y}_{<t}),\theta \right) 
\end{eqnarray}
\end{small}
In this way, MNMT minimizes the probability of predicting off-target tokens that are conditioned on the off-target input tokens of the decoder.
The objective function of unlikelihood training can be expressed as follows:
\begin{small}
\begin{eqnarray}
\label{eq:unlikelihood training framework}
\mathcal{F}_{UL} = \mathop{\arg\min} \sum_{l=l_1}^{l_N}\mathcal{L}_{UL} \left(X^{l}, \tilde{Y}^{En}, \theta \right) \!+\! \mathcal{L}_{UL} \left(X^{En}, \tilde{Y}^{l}, \theta \right)  \nonumber
\end{eqnarray}
\end{small}

\subsubsection{Model Selection Indicator}
\label{sec:Indicator}
As access to ZST samples is not permitted during training, we cannot directly select the final model according to the loss scores obtained on the dev set. Therefore, we select the final model based on the CWR separation degree in the zero-shot off-target setting.
Given the CWRs $P^{l_i}$ of a target language $l_i$ with a mean of $P_{mean}^{l_i}$ and a distance metric of $\mathbf{Dis}()$ between the two distributions, the separation degree is computed as follows:
\begin{equation}\label{eq:COWRIE}
\small
\mathcal{S}ep = \frac{\sum_{i \in N} \sum_{j \in N, j > i}  \mathbf{Dis} \left ( P^{l_i}, P^{l_j} \right ) * 2 }{\sum_{i \in N} \mathbf{Dis} \left ( P^{l_i}, P_{mean}^{l_i} \right ) * \left ( N + 1 \right )},
\end{equation}
where $N$ is the number of languages in $L$, where we set $\mathbf{Dis}(\cdot, \cdot)$ as the average distance between the CWR points of the different distributions.

As training proceeds, the separation degree increases, and we select the final model with converged scores. Specifically, we regard a model with changes below 0.01 as converged, and for efficient computation purposes, we only use 100 training samples and consider the languages included in the ZST test set.

\begin{table*}[t]
\centering
\caption{{\bf ZST performance achieved on the IWSLT-4 dataset.} `$\Delta$': improvements over the vanilla model. \underline{Underline}: the averaged results. \textbf{Bold}: the best results. `$^\ddagger$': statistically significant improvement ($p<0.01$).}
\label{tab:main_iwslt-4}
\small
\setlength{\tabcolsep}{5.3mm}{
\begin{tabular}{lccccccc}
\toprule
& \multicolumn{7}{c}{\bf IWSLT-4} \\ \hline \hline 
\multirow{2}{*}{\textbf{Model}} & \multicolumn{2}{c}{\textbf{Nl-Ro}}& \multicolumn{2}{c}{\textbf{Nl-It}}& \multicolumn{2}{c}{\textbf{It-Ro}}& \multicolumn{1}{c}{\multirow{2}{*}{\textbf{AVG}}} \\
 & \multicolumn{1}{c}{\textbf{$\leftarrow$}} & \multicolumn{1}{c}{\textbf{$\rightarrow$}} & \multicolumn{1}{c}{\textbf{$\leftarrow$}} & \multicolumn{1}{c}{\textbf{$\rightarrow$}} & \multicolumn{1}{c}{\textbf{$\leftarrow$}} & \multicolumn{1}{c}{\textbf{$\rightarrow$}} & \multicolumn{1}{c}{} \\ \hline
 & \multicolumn{7}{c}{\textbf{\textit{SacreBLEU Scores} $\uparrow$}} \\
\textbf{Vanilla} & 12.2 & ~~9.7  & 12.9 & 11.9 & 13.1 & 11.3 & \underline{11.9}\\
\bf \quad+UNIONS  & ~~\textbf{15.6}$^\ddagger$ & ~~\textbf{13.0}$^\ddagger$ & ~~\textbf{16.1}$^\ddagger$ & ~~\textbf{17.4}$^\ddagger$ & ~~\textbf{18.1}$^\ddagger$ & ~~\textbf{15.1}$^\ddagger$ & \textbf{\underline{15.9}}\\ 
 \quad\textbf{$\Delta$}& +3.4 & +3.3 & +3.2 & +5.5 & +5.0 & +3.8 & \underline{+4.0} \\ \hline
 & \multicolumn{7}{c}{\textbf{\textit{OTR Scores \%} $\downarrow$}} \\
\textbf{Reference}  & ~~2.8  & ~~4.3  & ~~2.9  & ~~0.7  & ~~0.8  & ~~4.4  & ~~\underline{2.7} \\
\cdashline{1-7}
\textbf{Vanilla} & 12.6 & 21.7 & 12.0 & 25.0 & 20.6 & 23.0 & \underline{19.2} \\
\bf \quad+UNIONS  & ~~\textbf{2.4}   & ~~\textbf{3.8}   & ~~\textbf{2.7}  & ~~\textbf{0.8}   & ~~\textbf{1.1}   & ~~\textbf{4.2}   & ~~\underline{\textbf{2.5}} \\
 \quad\textbf{$\Delta$}& -10.2~ & -17.9~ & ~-9.3 & -24.2~ & -19.5~ & -18.8~ & \underline{-16.7}~ \\
\bottomrule
\end{tabular}}
\end{table*}

\begin{table*}[t]
\centering
\caption{{\textbf{ZST performance achieved on the OPUS-100 (v1.0) dataset.}} ``$\Delta$'': improvements over the vanilla model. \underline{Underline}: the averaged results. \textbf{Bold}: the best results. `$^\ddagger$': statistically significant improvement ($p<0.01$).}
\label{tab:main_opus-100}
\resizebox{\linewidth}{!}{
\begin{tabular}{lccccccccccccc}
\toprule
& \multicolumn{13}{c}{\bf OPUS-100 (v1.0)} \\ \hline \hline 
\multirow{2}{*}{\textbf{Model}} & \multicolumn{2}{c}{\textbf{Fr-De}} & \multicolumn{2}{c}{\textbf{Ru-Fr}} & \multicolumn{2}{c}{\textbf{Nl-De}} & \multicolumn{2}{c}{\textbf{Zh-Ru}} & \multicolumn{2}{c}{\textbf{Zh-Ar}} & \multicolumn{2}{c}{\textbf{Nl-Ar}}& \multirow{2}{*}{\textbf{AVG}} \\
& \textbf{$\leftarrow$}    & \textbf{$\rightarrow$}   & \textbf{$\leftarrow$}    & \textbf{$\rightarrow$}   & \textbf{$\leftarrow$}    & \textbf{$\rightarrow$}   & \textbf{$\leftarrow$}    & \textbf{$\rightarrow$}   & \textbf{$\leftarrow$}    & \textbf{$\rightarrow$}   & \textbf{$\leftarrow$}   & \textbf{$\rightarrow$}  &\\ \hline
& \multicolumn{13}{c}{\textbf{\textit{SacreBLEU Scores} $\uparrow$}} \\
\textbf{TLP\&TGP}~\cite{yang-etal-2021-improving-multilingual}   & ~~6.6     & \textbf{14.2}    & 16.7    & \textbf{21.4}    & \textbf{16.2}    & ~~8.6     & 12.9    & 14.2    & 14.7    & \textbf{12.6}    &\textbf{11.8}   & ~~\textbf{4.6}    & \underline{12.9}  \\\cdashline{1-13}
\textbf{Vanilla}& ~~3.3  & ~~3.0  & ~~4.8  & ~~5.4  & ~~4.7  & ~~4.3  & ~~3.7  & ~~5.1  & ~~4.4  & ~~5.4  & ~~1.4 & ~~0.9 & ~~\underline{3.9}\\
\bf \quad+UNIONS      & ~~\textbf{14.9}$^\ddagger$  & ~~12.3$^\ddagger$ & ~~\textbf{17.1}$^\ddagger$  & ~~19.1$^\ddagger$  & ~~16.0$^\ddagger$  & ~~\textbf{15.2}$^\ddagger$  & ~~\textbf{23.0}$^\ddagger$  & ~~\textbf{14.5}$^\ddagger$ & ~~\textbf{25.2}$^\ddagger$  & ~~11.6$^\ddagger$ & ~~~~8.8$^\ddagger$  & ~~1.8  & \underline{\textbf{15.0}} \\ 
 \quad \textbf{$\Delta$}& +11.6~~ & +9.3 & +12.3~~ & +13.7~~ & +11.3~~ & +10.9~~ & +19.3~~ & +9.4 & +20.8~~ & +6.2 & +7.4 & +0.9 & \underline{+11.1}~~ \\ \hline
& \multicolumn{13}{c}{\textbf{\textit{OTR Scores \%} $\downarrow$}}\\
\textbf{Reference}& ~~4.7     & ~~3.3     & ~~2.9     & ~~4.5     & ~~7.1     & ~~3.4     & ~~6.0     & ~~3.8     & ~~6.4     & ~~4.1     & 10.0   & ~~2.3    & ~~\underline{4.9} \\
\textbf{TLP\&TGP}~\cite{yang-etal-2021-improving-multilingual}     & -       & -       & -       & -       & -       & -       & -       & -       & -       & -       & -      & -      & \underline{16.9} \\
\cdashline{1-13}
\textbf{Vanilla}& 89.8    & 98.7    & 96.2    & 92.5    & 98.0    & 97.4    & 94.6    & 96.0    & 93.5    & 85.2    & 98.4   & 84.8   & \underline{93.7} \\
\bf \quad+UNIONS & ~~\textbf{7.9}   & \textbf{19.9}  & ~~\textbf{8.8}   & ~~\textbf{8.0}   & \textbf{14.0}  & \textbf{11.2}  & \textbf{29.6}  & ~~\textbf{7.9}   & \textbf{21.9}  & ~~\textbf{5.6}   & \textbf{27.3}  & \textbf{12.8}  & \underline{\textbf{14.6}} \\ 
 \quad \textbf{$\Delta$}& -81.9~ & -78.8~ & -87.5~ & -84.5~ & -83.9~ & -86.2~ & -65.0~ & -88.2~ & -71.7~ & -79.6~ & -71.0~ & -71.9~ & \underline{-79.2}~   \\
\bottomrule
\end{tabular}}
\end{table*}

\begin{table*}[t]
  \centering
  \caption{{\textbf{ZST performance achieved on the WMT-5 dataset.}} ``$\Delta$'': improvements over the vanilla model. \underline{Underline}: the averaged results. \textbf{Bold}: the best results. `$^\ddagger$': statistically significant improvement ($p<0.01$).}
  \label{tab:main_wmt-5}
  \resizebox{\linewidth}{!}{
  \begin{tabular}{lccccccccccccc}
  \toprule
  & \multicolumn{13}{c}{\bf WMT-5} \\ \hline \hline 
  \multirow{2}{*}{\textbf{Model}} & \multicolumn{2}{c}{\textbf{Fr-De}}& \multicolumn{2}{c}{\textbf{Zh-De}}& \multicolumn{2}{c}{\textbf{Ro-De}}& \multicolumn{2}{c}{\textbf{Zh-Fr}}& \multicolumn{2}{c}{\textbf{Ro-Fr}}& \multicolumn{2}{c}{\textbf{Ro-Zh}}& \multirow{2}{*}{\textbf{AVG}} \\ 
  & \textbf{$\leftarrow$} & \textbf{$\rightarrow$} & \textbf{$\leftarrow$} & \textbf{$\rightarrow$} & \textbf{$\leftarrow$} & \textbf{$\rightarrow$} & \textbf{$\leftarrow$} & \textbf{$\rightarrow$} & \textbf{$\leftarrow$} & \textbf{$\rightarrow$} & \textbf{$\leftarrow$} & \textbf{$\rightarrow$} & \\  \hline
  & \multicolumn{13}{c}{\bf \textit{SacreBLEU Scores} $\uparrow$ }   \\ 
  \textbf{Vanilla}& 10.8 & ~~5.2 & 19.3 & ~~1.9 & ~~5.8 & ~~5.0 & 15.3 & ~~1.6 & ~~5.3 & ~~6.4 & ~~2.8 & 11.4 & ~~\underline{7.6}  \\
  \bf \quad+UNIONS &  ~~\textbf{29.6}$^\ddagger$  & ~~\textbf{22.0}$^\ddagger$  & ~~\textbf{27.0}$^\ddagger$ & ~~\textbf{14.1}$^\ddagger$  & ~~~~\textbf{8.5}$^\ddagger$  & ~~~~\textbf{7.8}$^\ddagger$  & ~~\textbf{27.2}$^\ddagger$  & ~~\textbf{19.7}$^\ddagger$  & ~~\textbf{10.3}$^\ddagger$ & ~~\textbf{16.2}$^\ddagger$ & ~~~~\textbf{5.5}$^\ddagger$  & ~~\textbf{13.0}$^\ddagger$ & \underline{\textbf{16.7}} \\ 
   \quad \bf$\Delta$ & +18.8~~ & +16.8~~ & +7.7 & +12.2~~ & +2.7 & +2.8 & +11.9~~ & +18.1~~ & +5.0 & +9.8 & +2.7 & +1.6 & \underline{+9.1} \\   \hline
   & \multicolumn{13}{c}{\bf \textit{OTR Scores \%} $\downarrow$ } \\
  \textbf{Reference} &  ~~0.0 & ~~0.0 & ~~0.0 & ~~0.0 & ~~0.0 & ~~0.0 & ~~0.0 & ~~0.0 & ~~0.0 & ~~0.0 & ~~0.0 & ~~0.0 & ~~\underline{0.0} \\ 
  \cdashline{1-13}
  \textbf{Vanilla}& 69.1  & 84.5  & 23.3  & 98.1  & 50.3  & 54.5  & 39.1  & 99.5  & 63.1  & 68.6  & 76.5  & 19.9  & \underline{62.2} \\
  \bf \quad+UNIONS &  ~~\textbf{1.0}   & ~~\textbf{1.0}   & ~~\textbf{1.2}   & ~~\textbf{0.4}   & ~~\textbf{3.5}   & ~~\textbf{1.5}   & ~~\textbf{1.1}   & ~~\textbf{1.0}   & ~~\textbf{2.8}   & ~~\textbf{4.1}   & ~~\textbf{2.9}   & ~~\textbf{5.8}   & ~~\underline{\textbf{2.2}} \\ 
   \quad \bf$\Delta$ & -68.1~ & -83.5~ & -22.1~ & -97.7~ & -46.8~ & -53.0~ & -38.0~ & -98.5~ & -60.3~ & -64.5~ & -73.6~ & -14.1~ & \underline{-60.0}~  \\  
  \bottomrule
  \end{tabular}}
  \end{table*}

\begin{table*}[t]
\centering
\caption{{\bf ZST performance achieved on the TED dataset.} `$\Delta$': improvements over the vanilla model. \underline{Underline}: the averaged results. \textbf{Bold}: the best results. `$^\ddagger$': statistically significant improvement ($p<0.01$). $l_{mono}$ only has monolingual data during training, while $l_{bi}$ exists in the multilingual translation dataset. }
\label{tab:main_adapter}
\resizebox{\linewidth}{!}{
\begin{tabular}{lccccccccccc}
\toprule
& \multicolumn{10}{c}{\textbf{TED}} & \\  \hline \hline 
\multirow{2}{*}{\textbf{Models}} & \multicolumn{4}{c}{\textbf{$l_{bi} \rightarrow l_{bi}$}} & \multicolumn{2}{c}{\textbf{$l_{mono} \rightarrow l_{mono}$}} & \multicolumn{2}{c}{\textbf{$l_{bi} \rightarrow l_{mono}$}} & \multicolumn{2}{c}{\textbf{$l_{mono} \rightarrow l_{bi}$}} & \multirow{2}{*}{\textbf{AVG}} \\ 
& Ko$\rightarrow$ It  & Zh$\rightarrow$ Fr & Pl$\rightarrow$ De & My$\rightarrow$ Hi & Es$\rightarrow$ Nl  & Lt$\rightarrow$ Et  & Zh$\rightarrow$ Nl& Pl$\rightarrow$ Et  & Uk$\rightarrow$ Cs & Fi$\rightarrow$ De & \\ \hline
& \multicolumn{10}{c}{\textbf{\textit{SacreBLEU Scores} $\uparrow$}} &  \\ 
\textbf{D.A.}~\cite{ustun-etal-2021-multilingual}     & ~~5.5  & ~~9.0 & ~~8.2 & ~~1.0 & ~~7.2  & ~~3.6  & ~~4.1 & ~~4.9 & ~~9.2 & ~~9.6 & ~~\underline{6.2}\\
\bf \quad+UNIONS  & ~~\textbf{6.3} & \textbf{10.3}$^\ddagger$ & ~~\textbf{9.0} & ~~\textbf{1.2} & \textbf{11.4}$^\ddagger$ & ~~\textbf{4.4} & ~~\textbf{5.3}$^\ddagger$  & ~~\textbf{5.8} & ~~\textbf{9.9} & \textbf{11.6}$^\ddagger$ & ~~\underline{\textbf{7.5}}\\
 \quad \textbf{$\Delta$}  & +0.8  & +1.3 & +0.8 & +0.2 & +4.2  & +0.8  & +1.2 & +0.9 & +0.7 & +2.0 & \underline{+1.3}\\ \hline
& \multicolumn{10}{c}{\textbf{\textit{OTR Scores \%} $\downarrow$}} & \\
\textbf{Reference} & ~~1.3  & ~~0.5 & ~~1.1 & ~~1.2 & ~~3.5  & ~~8.1  & ~~3.4 & ~~6.2 & ~~2.6 & ~~0.9 & ~~\underline{2.9}\\ \hdashline
\textbf{D.A.}~\cite{ustun-etal-2021-multilingual}      & 20.2 & 12.2& 21.1& 14.7& 46.3 & 24.3 & 40.5  & 23.7& 11.2& 18.3& \underline{23.2}  \\
\bf \quad+UNIONS   & ~~\textbf{5.3}  & ~~\textbf{4.1} & ~~\textbf{7.2} & ~~\textbf{7.6} & \textbf{15.4} & ~~\textbf{8.1}  & \textbf{14.7 } & \textbf{10.9 }& ~\textbf{~6.1} & ~~\textbf{4.8} & ~~\underline{\textbf{8.4}}\\
 \quad \textbf{$\Delta$}  & -14.9~ & -8.1 & -13.9~ & -7.1 & -31.0~ & -16.2~ & -25.8~ & -12.8~ & -5.1 & -13.4~ & \underline{-14.8}~ \\
\bottomrule
\end{tabular}}
\end{table*}

\section{Task Setup}
\label{sec: task}
We evaluate UNIONS on ZST tasks spanning 40 directions and two types of base models to verify the effectiveness and universality of our approach.
\subsection{Multilingual Machine Translation Model}
\label{sec:RI_settings}

We begin by examining models trained on English-centric multilingual translation datasets and proceed to evaluate these models directly on extensive zero-shot translations (e.g., non-English translations XX$\leftrightarrow$XX).
We take the following three translation benchmarks into consideration.
\begin{itemize} 
    \item \textbf{IWSLT-4}: Following \citet{qu-watanabe-2022-adapting}, we use the IWSLT-17 dataset to evaluate the performance of the models, and we remove four languages ( ``En, Ro, It, Nl'') from MMCR4NLP~\cite{dabre2017mmcr4nlp}. IWSLT-4 is a multialigned dataset with 145k training sentences for each language.
    \item \textbf{OPUS-100 (V1.0)}: We also conduct experiments on the OPUS-100 (v1.0) dataset from \citet{yang-etal-2021-improving-multilingual}. OPUS-100~\cite{zhang-etal-2020-improving} is an English-centric dataset that has 55 M samples with a maximum of 1 M sentence pairs for each language pair. It consists of parallel corpora between En and 100 other languages. Following \citet{yang-etal-2021-improving-multilingual}, we construct OPUS-100 (v1.0) by removing 5 languages (``An, Dz, Hy, Mn, Yo'') without training or testing data and removing all duplicate test sentence pairs from the training and testing sets.
    \item \textbf{WMT-5}: To further evaluate the extremely unbalanced large-scale scenario, we adopt 4 popular WMT parallel training datasets, including WMT14 En$\leftrightarrow$De (4.5 M), WMT14 En$\leftrightarrow$Fr (35.7 M), WMT16 En$\leftrightarrow$Ro (608 k) and WMT17 En$\leftrightarrow$Zh (20.5 M). To prevent language cover bias during the evaluation, we use the multialigned Flores-200 devtest set\footnote{\url{https://github.com/facebookresearch/flores/blob/main/flores200/README.md}}~\cite{costa2022no} for all translations. 
\end{itemize}

We compare our model with two strong baselines, including vanilla and TLP$\&$TGP:
\begin{itemize} 
    \item \textbf{Vanilla}: We use the vanilla MNMT model by closely following the optimal model and training settings of~\citet{qu-watanabe-2022-adapting,yang-etal-2021-improving-multilingual} except for the language ID setting, where we follow \citet{pan-etal-2021-contrastive, arivazhagan2019missing} to prepend the source and target language IDs to the inputs of the encoder and decoder, respectively. In contrast, \citet{qu-watanabe-2022-adapting,yang-etal-2021-improving-multilingual} prepended the target ID into the encoder. Our model is further tuned by the vanilla method.
    \item \textbf{TLP$\&$TGP~\cite{yang-etal-2021-improving-multilingual}}: On the OPUS-100 (v1.0) dataset, we also compare the proposed method with TLP$\&$TGP, which regularize MNMT models at both the representation and gradient levels. We compare our results with those reported by \citet{yang-etal-2021-improving-multilingual}.
\end{itemize}

We conduct experiments on the {\tt fairseq}~\cite{ott2019fairseq} toolkit with a transformer~\cite{NIPS2017_3f5ee243} as the MNMT backbone and select the final model according to the loss induced on the dev sets.
To balance the distributions of different parallel corpora, we follow \citet{arivazhagan2019massively} and use a temperature-based sampling method and set $T=5$. 
We tokenize IWSLT-4 via a 40k vocabulary and split the words in the OPUS-100 (v1.0) and WMT-5 corpora into subword units using a 64k SentencePiece~\cite{kudo-richardson-2018-sentencepiece} vocabulary to train their corresponding training sets. 

For IWSLT-4, we use a 5-layer transformer with 8 attention heads, an embedding size of 512, an inner size of 2048, a training process with a dropout of 0.3, an lr of 5e-4, a 16k batch size, a label smoothing parameter of 0.1, and 100k update steps. In experiments conducted on the large-scale OPUS-100 (v1.0) and WMT-5 datasets, we use Transformer-big with 6 layers and 16 attention heads and set the warmup parameter to 4k, the batch size to 524k, the label smoothing parameter to 0.1, the dropout parameter to 0.1, and the attention dropout parameter to 0.1. In addition, we set the lr to 5e-4 for OPUS-100 (v1.0) and to 7e-4 for WMT-5. 

For fine-tuning, we set the warmup parameter to 1, the lr to 5e-5, the batch size to 1k, and the number of updates to 0.5k for IWSLT-4; we set the lr to 7e-5, the batch size to 32k, and the number of updates to 0.5k for WMT-5; and we set the lr to 5e-5, the batch size to 32k, and the number of updates to 5k for OPUS-100 (v1.0). 
For evaluation purposes, we save checkpoints every 50 updates for IWSLT-4/WMT-5 and every 500 updates for OPUS-100 (v1.0). We choose the final model according to the indicator defined in \S\ref{sec:Indicator}. 

All models are trained on Tesla-A100 GPU. Note that to conduct a fair comparison with Vanilla, we report the results obtained with a continued training setting that is identical to the tuning steps of our UNIONS method.
We adopt \textbf{SacreBLEU}~\cite{post-2018-call} to evaluate the translation accuracy, where we generate translations with a beam size of 5. We also compute the off-target ratio of the generated \textbf{OTR scores} with a publicly available language detector\footnote{\url{https://fasttext.cc/docs/en/language-identification.html}}~\cite{joulin2016fasttext,joulin-etal-2017-bag}. These two evaluation metrics are also used in adapter tuned translation model experiments.

\subsection{Adapter Tuned Translation Model}
Adapter tuning~\cite{pmlr-v97-houlsby19a,he2022sparseadapter,he2023mera} adapts large pretrained language models (PLMs) for downstream tasks by incorporating lightweight residual layers into each model layer. During training, the adapter layers are fine-tuned using downstream data while the other parameters remain frozen.
We take the open source \textbf{denoising adapters (D.A.) model}\footnote{\url{https://europe.naverlabs.com/research/natural-language-processing/efficient-multilingual-machine-translation-2}}~\cite{ustun-etal-2021-multilingual} as our base model and tune it with UNIONS, similar to the MNMT models in \S\ref{sec:RI_settings}.
D.A. is a two-stage tuned model based on mBART50~\cite{tang-etal-2021-multilingual}. It contains 1) training adapters for each language with a denoising task for 37 languages\footnote{Languages with monolingual data: Ar, He, Ru, Ko, It, Ja, Zh, Fr, Pt, Tr, Ro, Pl, Vi, De, Fa, Cs, Th, My, Hi, Es, Nl, Hr, Uk, Id, Sv, Lt, Fi, Et, Ur, Kk, Bg, Hu, Sr, El, Da, and Be. } containing monolingual data with a maximum of 20 million sentences per language and 2) cross-attention trained on the TED talks~\cite{qi-etal-2018-pre} dataset by selecting 20 languages\footnote{Languages with En$\leftrightarrow$XX parallel data: Ar, He, Ru, Ko, It, Ja, Zh, Fr, Pt, Tr, Ro, Pl, Vi, De, Fa, Cs, Th, My, and Hi.} with training sizes ranging from 214k to 18k parallel sentences. During translation, the target language ID and the corresponding adapter determine the translation flow. While D.A. only evaluates translation performance with English as the reference language, we further enhance its translation ability for non-English languages.

During the tuning process of UNIONS, we use the same TED talk dataset and set 1 warmup step, an lr of 1e-5, 1024 max tokens, and 1K updates.
We save a checkpoint every 100 updates. The evaluation process remains the same as that used in MNMT.

\begin{table*}[t]
\centering
\caption{\textbf{Supervised translation performance comparison.} ``$\Delta$'': improvements over the vanilla model. \textbf{Bold}: the best results. \underline{Underline}: average scores obtained for all supervised directions.}
\label{tab:supervised}
\resizebox{\linewidth}{!}{
\begin{tabular}{lccccccccc}
\toprule
\multirow{2}{*}{\textbf{Model}} & \multicolumn{3}{c}{\textbf{IWSLT-4}}    & \multicolumn{3}{c}{\textbf{OPUS-100 (v1.0)}}   & \multicolumn{3}{c}{\textbf{WMT-5}} \\ \cmidrule(r){2-4} \cmidrule(r){5-7} \cmidrule(r){8-10}
& \bf En$\rightarrow$XX & \bf  XX$\rightarrow$En & \bf  AVG & \bf  En$\rightarrow$XX & \bf  XX$\rightarrow$En & \bf  AVG & \bf  En$\rightarrow$XX & \bf  XX$\rightarrow$En & \bf  AVG \\ \hline
\textbf{Vanilla}    & 27.8  & 31.8  & \underline{29.8} & 25.7 & 32.3 & \underline{29.0} & 33.1  & \textbf{31.7} & \underline{\textbf{32.4}} \\
\bf \quad+UNIONS & \textbf{27.9} & \textbf{32.0} & \underline{\textbf{29.9}} & \textbf{25.8} & \textbf{32.7} & \underline{\textbf{29.2}} & \textbf{33.2}  & 31.4 & \underline{32.3} \\ 
 \quad \textbf{$\Delta$}   &  +0.1 & +0.2  & \underline{+0.1} & +0.1 & +0.4 &\underline{+0.2}~& +0.1 & ~-0.3 & ~\underline{-0.1} \\
\bottomrule
\end{tabular}}
\end{table*}

\section{Experiments}
\label{sec: main results}
In this section, we conduct extensive experiments spanning 40 ZST directions to verify the effectiveness and universality of our UNIONS method.

\subsection{UNIONS Achieves Considerably Improved ZST Performance}
The main results obtained on the three benchmarks in Tables \ref{tab:main_iwslt-4}, \ref{tab:main_opus-100} and \ref{tab:main_wmt-5} show that our method achieves consistent and significant improvements over the vanilla model for both large- and small-scale datasets, as well as with different model sizes (Transformer-small is used for IWSLT, while Transformer-big is used for OPUS-100 (v1.0) and WMT-5).
In particular, our method effectively reduces the off-target ratios compared to those of the baseline by $-16.7\%/-79.2\%/-60.0\%$ for the IWSLT, OPUS, and WMT-5 benchmarks, respectively, thus significantly improving the resulting translation quality by $+4.0/+14.6/+9.1$ averaged SacreBLEU score points.

Additionally, we observed that MNMT trained on large-scale datasets, such as OPUS-100 and WMT-5, face more severe off-target problems.
And, UNIONS achieves the greatest improvement on the difficult OPUS-100 (v1.0) benchmarks with the largest number of languages.
For IWSLT-4, the improvement is minimal, which we attribute to the lower upper bound of an MNMT model trained on a low-resource dataset.
In comparison with previous works utilizing OPUS, our model outperforms TLP$\&$TG by $+2.1$ SacreBLEU and $-2.0\%$ OTR scores.

\subsection{UNIONS Provides Greater Benefits for Target Languages that are Close to English}
We consider the influence of the similarity between languages on the performance of the model.
According to the OTR scores obtained on OPUS-100 (v1.0) and WMT-5 and shown in Tables~\ref{tab:main_opus-100} and \ref{tab:main_wmt-5}, we find that our method is particularly effective for ZST tasks with target languages that are similar to English, e.g., Zh$\rightarrow$Ru ($-88.2\%$) in Table~\ref{tab:main_opus-100} and Zh$\rightarrow$De ($-97.7\%$), Zh$\rightarrow$Fr ($-98.5\%$), and Zh$\rightarrow$Ro ($-73.6\%$) in Table~\ref{tab:main_wmt-5}. These languages are all in the Indo-European language family. 
In contrast, for their reverse directions, where the target languages are in non-English families, the achieved improvements are smaller, e.g., Ru-Zh ($\Delta23.2\%$) in Table~\ref{tab:main_opus-100} and Zh-De ($\Delta75.6\%$), Fr-Zh ($\Delta60.5\%$), and Ro-Zh ($\Delta59.5\%$) in Table~\ref{tab:main_wmt-5}, where $\Delta$ represents the gap of between the OTR scores obtained in these directions and those obtained in the forward directions.

We attribute this interesting phenomenon to the fact that our method is able to significantly enhance the navigation capabilities of the IDs of languages that are close to the central language (English), which are weaker in the vanilla model.

\subsection{Adapter-Tuned Translation Model}
As reported in Table~\ref{tab:main_adapter}, our method achieves an average SacreBLEU score improvement of +1.3 and reduces the off-target ratio by -14.8\% compared to that of the baseline D.A. model. These results demonstrate that UNIONS also effectively addresses off-target problems in fine-tuned translation models based on PLMs.

We utilize $l_{mono}$ and $l_{bi}$ to indicate whether the current language has bitext data during the D.A. training process. For instance, in the zh$\rightarrow$nl case, we use $ l_{bi} \rightarrow l_{mono}$ to denote that the training set of D.A. has zh$\rightarrow$en bitext data and $nl$ monolingual data.
We find that UNIONS is more effective for target languages that only have monolingual data, e.g., Es$\rightarrow$Nl (-31.0\% OTR score), Lt$\rightarrow$Et (-16.2\% OTR score), Zh$\rightarrow$Nl (-25.8\% OTR score) and Pl$\rightarrow$Et (-12.8\% OTR score).
One possible explanation for this observation is that the language IDs, which are trained solely on self-supervised objectives using monolingual data, have more delicate navigation capabilities, making them more susceptible to improvement through the UNIONS approach.

\section{Analysis}
To provide some insights to better understand our proposed method, i.e., UNIONS, we conduct extensive analyses from different perspectives to show 1) the maintained supervised translation performance, 2) the effectiveness of our model selection approach, 3) the negligible computational cost, and 4) the rejuvenated navigation capability of language IDs.
\label{sec: analysis}

\subsection{UNIONS Maintains the Supervised Translation Performance of the Model}
As our approach aims to optimize the zero-shot performance of a pretrained MNMT model, one may doubt whether the supervised translation performance is affected. To dispel this concern, we report the averaged SacreBLEU scores obtained for the supervised directions, including translating from English (En $\rightarrow$ XX) and translating into English (XX $\rightarrow$ En), on three benchmarks in Table~\ref{tab:supervised}. 

UNIONS demonstrates improvements of +0.1 and +0.2 average SacreBLEU scores over the vanilla baseline in IWSLT-4 and OPUS-100 (v1.0) respectively. And, the average performance drop of WMT-5 is negligible (-0.1 SacreBLEU score).
Overall, UNIONS achieves comparable performance to that of the vanilla MNMT model for all benchmarks (with an average translation performance improvement of +0.1 SacreBLEU score), demonstrating that \textbf{\em our model successfully maintains its supervised translation ability}.

\subsection{Effectiveness of Our Model Selection Approach During Training}
\begin{figure}
     \centering
    \includegraphics[width=0.50\textwidth]{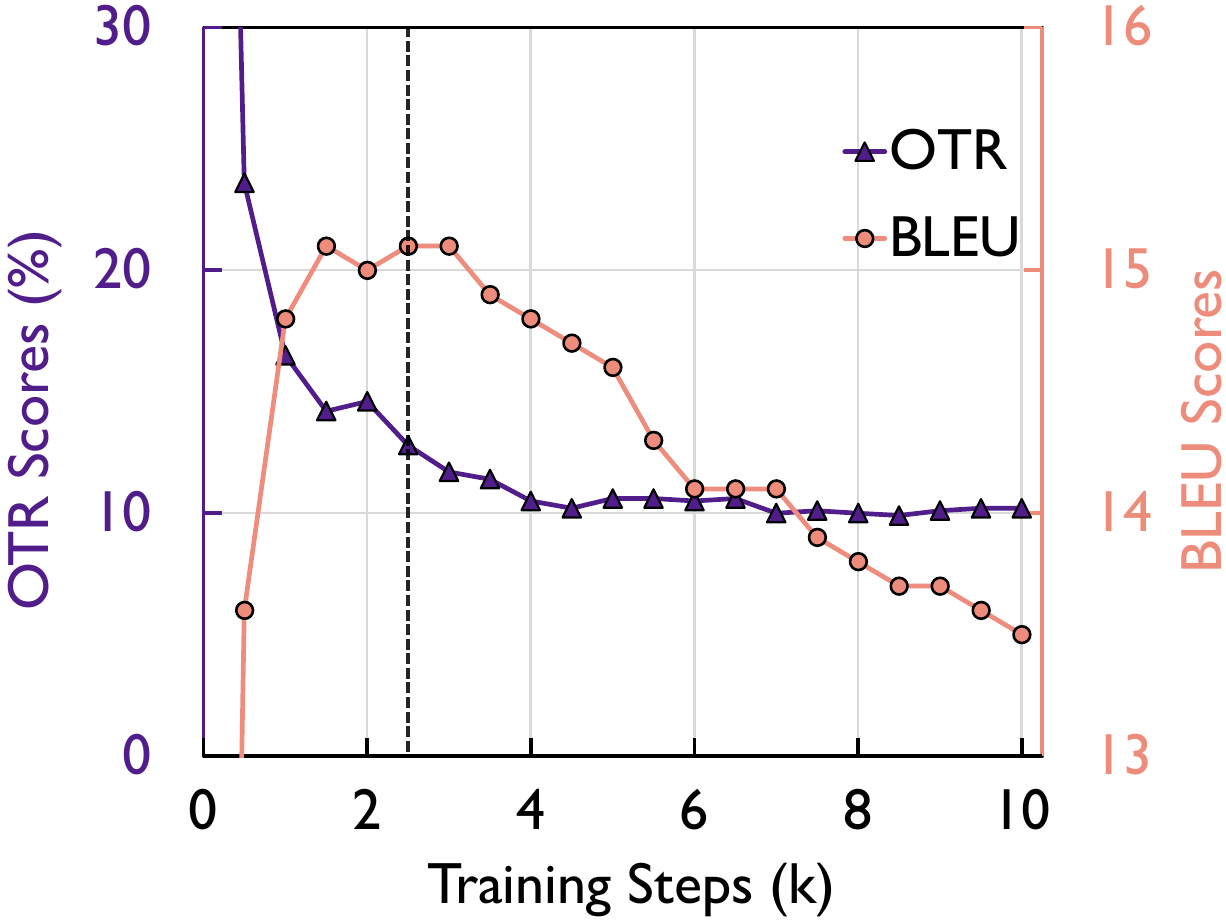}
     \caption{\textbf{Training dynamics exhibited by UNIONS on OPUS.} We present the SacreBLEU and OTR scores according to the number of training steps. The dashed line indicates the final model selected by $\mathcal{S}ep$ in \S\ref{sec:Indicator}.}
    \label{fig:step_abalation}
\end{figure}

As mentioned above, the CWR separation degree in \S\ref{sec:Indicator} is used to select the checkpoints.
To validate the effectiveness of our approach, we tune a trained MNMT model on OPUS-100 (v1.0) with UNIONS for 10K steps and report the SacreBLEU and OTR scores produced on zero-shot test sets during the training process in Figure~\ref{fig:step_abalation}.

As seen, 1) indicator $\mathcal{S}ep$ can easily choose the best checkpoint in 2.5K out of 10K steps, where the model has decent translation performance and a relatively low OTR score, showing the \textbf{\em effectiveness of our proposed proxy model selection indicator} in \S\ref{sec:Indicator}; 2) the dynamics of the OTR scores exhibit a significant decline first and then stabilize, demonstrating the effectiveness of reducing the off-target ratios; and 3) the BLEU dynamics yield higher results in the early stages (approximately 2K), then gradually decrease after 3K, and finally still exceed those of the untuned MNMT model. This may be due to the overfitting of the unlikelihood loss on negative samples, and this feature should be explored to stabilize the learning process in future work. Luckily, our indicator selects a good checkpoint for achieving better OTR and BLEU scores before the unstable learning dynamic appears.

\subsection{UNIONS Requires a Negligible Computational Cost}

\begin{table}[t]
\centering
\caption{\textbf{GPU hours of model training.} ``Vanilla MNMT'' refers to the cost of training the MNMT model, ``+UNIONS'' represents the tuning cost of our model, and ``Ratio'' is calculated as the cost of "+UNIONS" divided by the cost of ``Vanilla MNMT'' (\%).}
\label{tab:cost}
\resizebox{0.99\linewidth}{!}{
\begin{tabular}{lcc}
\toprule
\textbf{Stage} & \bf OPUS-100 (v1.0) & \bf WMT-5 \\ \midrule
\bf Vanilla MNMT & $\approx$139.6  & $\approx$276.1 \\
\bf \quad+UNIONS & $\approx$2.5~~~~  & $\approx$0.8~~~~ \\ \hdashline
\bf \quad Ratio & $\approx$1.8\%  & $\approx$0.3\% \\
\bottomrule
\end{tabular}}
\end{table}

The training cost is a critical factor for the practical implementation of neural network models~\cite{shen2023efficient}, and users often prefer methods that are both effective and efficient.
To evaluate the efficiency of UNIONS, we report the computation budget required for tuning models trained on large-scale datasets, including OPUS-100 (v1.0) and WMT-5. Specifically, we use the number of GPU hours to quantitatively compare UNIONS with the vanilla MNMT training process.

As shown in Table~\ref{tab:cost}, approximately 139.6 and 276.1 GPU hours are needed to train MNMT models on OPUS-100 (v1.0) and WMT-5, respectively. And, UNIONS requires 2.5 and 0.8 GPU hours to fine-tune the models, thereby reducing off-target ratios thus boosting ZST translation performance. Considering the significant improvements shown in Table~\ref{tab:main_opus-100} and Table~\ref{tab:main_wmt-5}, \textbf{\em the addition training cost ratio, i.e.,1.8\% for OPUS-100 (v1.0) and 0.3\% for WMT-5, of UNIONS is negligible.}

\subsection{The Navigation Capabilities of Language IDs in ZST Can be Rejuvenated}

To understand whether our method can rejuvenate the navigation capabilities of language IDs facing off-target tokens in ZST, we use our model and reconduct the visualization analysis in \S\ref{three_setting_for_vis} for comparison with the vanilla model.  

\begin{figure}
     \centering
     \includegraphics[width=0.50\textwidth]{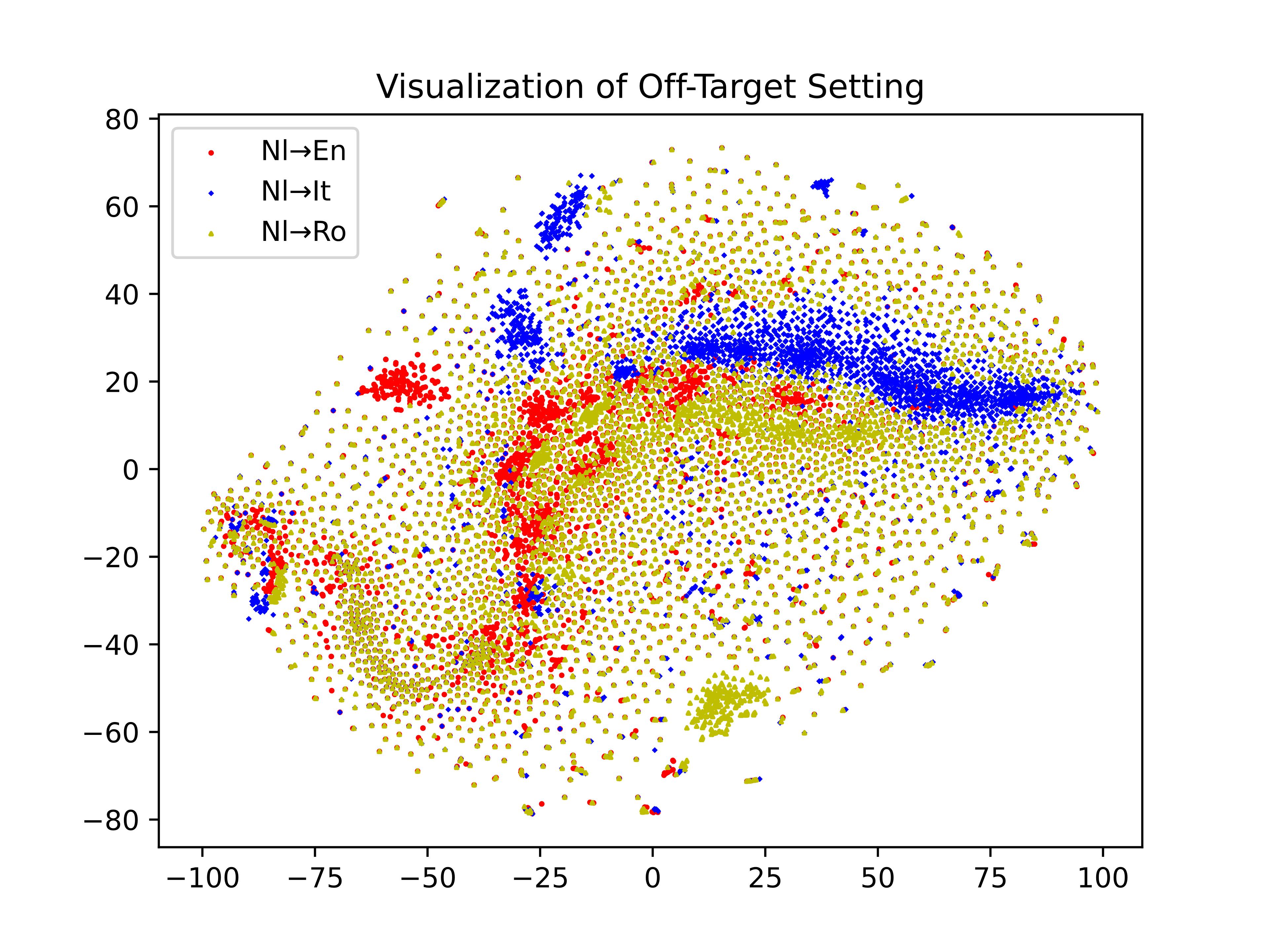}
     \caption{\textbf{Visualization of the CWRs generated by our model in the zero-shot off-target setting} in \S\ref{three_setting_for_vis}. Our model divides and navigates different languages into separate regions.}
    \label{fig:ours_cwr_vis}
\end{figure}

As depicted in Figure~\ref{fig:ours_cwr_vis}, in contrast with Figure~\ref{fig:cwr_vis}(d), the MNMT model clearly divides embeddings for different languages into separate regions under the zero-shot off-target setting after tuning with UNIONS.
This phenomenon is similar to the zero-shot on-target setting in Figure~\ref{fig:cwr_vis}(b) and the supervised settings in Figure~\ref{fig:cwr_vis}(a) and Figure~\ref{fig:cwr_vis}(c), where the model exhibits no confusion regarding the translation direction. 
The higher separate ratio is also consistent with the lower OTR score (2.5 v.s. 19.2) reported in Table~\ref{tab:main_iwslt-4}. 
This suggests that \textbf{\em UNIONS rejuvenates the lost navigation capabilities of language IDs}, confirming our claim.

\section{Conclusion}
\label{sec: conclusion}
In this paper, we first revisit the off-target problem in zero-shot machine translation (ZST) and show that language IDs are able to guide the translation flow while remaining fragile when faced with off-target tokens, which commonly exist during inference but are rare during training.
To address this issue, we propose a \textit{simple but sufficient} method -- UNIONS -- to minimize the probability of \textit{easy-to-construct} negative (language- and ID-unmatched) samples and bridge the gap between the MNMT training and ZST inference processes.
Our method possesses a simple training strategy that can improve any pretrained MNMT model: continued tuning with UNIONS.

Experimentally, UNIONS effectively reduces the off-target ratios in translation tasks and improves the resulting translation quality with a negligible extra computational cost.
Encouragingly, UNIONS provides more gains for larger-scale datasets, making it particularly beneficial for industry-level machine translation participants.

In future work, we will further investigate the impact of our method on non-English-centric data, e.g., parallel data between non-English languages. Meanwhile, it will be interesting to further design more effective methods to boost the zero-shot translation ability via post-training or tuning a pretrained generative language model, e.g., GPT-3~\cite{brown2020language}, BLOOM~\cite{scao2022bloom}, and LLaMA~\cite{touvron2023llama}.

\ifCLASSOPTIONcaptionsoff
  \newpage
\fi

\bibliographystyle{IEEEtranN}
\bibliography{UNIONS_arxiv}

\end{document}